\documentclass{article}

\usepackage{arxiv}
\usepackage{siunitx}
\usepackage{booktabs}
\usepackage{multirow}
\usepackage{algorithm}
\usepackage{algorithmic}
\usepackage{dsfont}
\usepackage{caption}
\usepackage{tabularx}
\usepackage[export]{adjustbox}
\usepackage{tikz}
\usepackage{amssymb}
\usepackage{amsmath}
\usepackage{hyperref}
\DeclareMathOperator*{\argmax}{arg\,max}
\DeclareMathOperator*{\argmin}{arg\,min}

\title{A Model-Based Reinforcement Learning Approach for a Rare Disease Diagnostic Task.}

\author{
  R\'emi~Besson \\
  CMAP\\
  \'Ecole Polytechnique\\
  Route de Saclay, 91128 Palaiseau \\
  \texttt{remi.besson@polytechnique.edu} \\
   \And
 Erwan~Le Pennec \\
   CMAP\\
  \'Ecole Polytechnique\\
  Route de Saclay, 91128 Palaiseau \\
  \texttt{erwan.le-pennec@polytechnique.edu} \\
    \And
 St\'ephanie~Allassonni\`ere \\
  School of Medicine\\
  Paris-Descartes University\\
  15 Rue de l'\'Ecole de M\'edecine, 75006 Paris \\
  \texttt{stephanie.allassonniere@parisdescartes.fr} \\
    \And
 Julien~Stirnemann \\
  Necker-Enfants Malades Hospital\\
  Paris-Descartes University\\
  149 Rue De S\`evres, 75015 Paris \\
  \texttt{julien.stirnemann@nck.aphp.fr} \\
    \And
  Emmanuel~Spaggiari\\
  Necker-Enfants Malades Hospital\\
  Paris-Descartes University\\
  149 Rue De S\`evres, 75015 Paris \\
  \texttt{emmanuel.spaggiari@aphp.fr} \\
    \And
 Antoine~Neuraz \\
  Department of Medical Informatics\\
  Necker-Enfants Malades Hospital\\
  149 Rue De S\`evres, 75015 Paris \\
  \texttt{antoine.neuraz@aphp.fr} \\
}

\begin{document}
\maketitle

\begin{abstract}
In this work, we present our various contributions to the objective of building a decision support tool for the diagnosis of rare diseases. Our goal is to achieve a state of knowledge where the uncertainty about the patient's disease is below a predetermined threshold. We aim to reach such states while minimizing the average number of medical tests to perform. In doing so, we take into account the need, in many medical applications, to avoid, as much as possible, any misdiagnosis. To solve this optimization task, we investigate several reinforcement learning algorithm and make them operable in our high-dimensional and reward-sparse setting. We also present a way to combine expert knowledge, expressed as conditional probabilities, with real clinical data. This is crucial because the scarcity of data in the field of rare diseases prevents any approach based solely on clinical data. Finally we show that it is possible to integrate the ontological information about symptoms while remaining in our probabilistic reasoning. It enables our decision support tool to process information given at different level of precision by the user. 

\end{abstract}

\newcommand{\enstq}[2]{\left\{#1~\middle|~#2\right\}}

\section{Introduction}

\subsection{Motivation.}
During pregnancy, several fetal ultrasounds are performed to evaluate the anatomy, the growth and the well-being of the fetus. During each ultrasound examination, the physician performs standardized measurements such as nuchal translucency and biometry as well as a predefined routine set of ultrasound planes of anatomical structures. Nevertheless in case of an anomaly, possibly related to a genetic disorder, there is no consensus on how to conduct the ultrasound in order to achieve the diagnosis of the disorder. It is a problem since there are many possible symptoms (around 200 in our case) that may be hard to detect and many more possible diagnoses while physicians do not have infinite time to check them all.

In this work, we want to systematize the prenatal diagnostic procedure in order to help the practitioner to make the diagnosis with high probability while minimizing the average number of questions, i.e symptom to check. To that purpose we design an algorithm that propose the most promising symptoms to check at each stage of the medical examination (state of knowledge about patient's condition, this kind of algorithm is sometimes called symptom checker in the literature) and provides the probability of each possible diseases. Eventually, our algorithm has to be operable and interpretable online, at the bedside.   

It should be noted that our approach applies to any problem aimed at establishing a symptom checker for rare diseases and is, of course, not limited to our case study to which we will refer throughout this paper: the prenatal diagnosis.

\subsection{Available data and some dimensions of the problem.}
\label{data}
The diseases we are interested in can be defined by a combination of symptoms, each with various degrees of likeliness. 
The data is structured as a list of diseases with their estimated probability and a list of symptoms for each disease (that we will call associated symptoms) with an estimation of the probability of the symptom given the disease.

We write:
$$S_i=\begin{cases}
1 &\text{if the fetus has the symptom of type i}\\
0 &\text{otherwise} \,.
\end{cases}$$
We denote the diseases: $D\in \{d_1,...,d_k\}$.
We know $P[D=d_j]=:P[D_j]$ as well as $P[S_i=1\mid D=d_j]=:P[S_i\mid D_j]$. Note that joint distribution of symptoms given the disease is not available, but only the marginals. This issue will be addressed in section \ref{learnmodel}. 

All this information, that we will refer to as expert data, has been provided by physicians of Necker hospital based on the available literature. We mapped all the symptoms found in the literature to the Human Phenotype Ontology (HPO) , see \cite{Khler2017TheHP}. HPO  is a recent work which provides a standardized vocabulary of phenotypic abnormalities encountered in human disease. We used it to harmonize the terminology. We could then combine our curated list of symptoms per disease and map it to OrphaData \footnote{Orphanet. INSERM 1997. An online rare disease and orphan drug data base. Available on http://www.orpha.net. Accessed [02/10/2018]}. OrphaData was useful to fill the missing data on prevalence of symptoms in the diseases. We restricted our analyses to the subset of symptoms that can be detected using fetal ultrasound. For these symptoms, we have extracted the information of the underlying tree structure ontology. We mean here by ontology the fact that a given symptom can be described at different level of precision: for example "\textit{heart deffect}" is an ascendant of "\textit{Tetralogy of Fallot}" (which is a specific cardiac abnormality). Our final decision support tool should handle such common medical reasoning (see section \ref{ontology} for more details).  

Currently, our database references $81$ diseases and $220$ different symptoms. The disease with the largest number of associated symptoms is VACTERL syndrome with $19$ possible symptoms.   

We will make the assumption that a fetus presents only one disease at a time which is a reasonable hypothesis in our case of rare disease study.

\subsection{Main contributions:}

In this work we present our different contributions to the objective of building a symptom checker for rare diseases. First we propose a novel notion, as fas as we know, of what should be a good symptom checker, taking into account the need in medicine to have a high level of confidence in the diagnosis made. This result in an original optimization formulation for a symptom checker building task. We found a way to break the dimension of our problem so as to make reinforcement learning algorithms tractable in this case. 

We also detail how to build an architecture drawing on both expert and clinical data so as to cope with a common issue in medicine (and even more so when it comes to rare diseases): the small amount of available clinical data.  

Finally we show that it is possible to incorporate the information of the symptoms ontology resulting in a much less rigid decision tool without computation explosion. 

All codes have been written in R language. In order to ensure reproducibility we made this code publicly available on GitHub.

\section{The sequential decision making problem: a planning task}
\label{planning}
\subsection{A Markov Decision Process framework}
\subsubsection{What we aim to optimize}

Our sequential decision problem can be formulated in the Markov Decision Process framework. Let $\mathbb{S}$ be the state space, using ternary base we encodes $1$ if the considered symptom is present, $0$ if it is absent, $2$ if non observed yet. We write $\mathbb{S}= \big\{(2,\dots,2),(1,2,\dots,2),\dots,(0,\dots,0)\big\}$. An element $s\in \mathbb{S}$ is a vector of length $220$ (the number of possible symptoms), it sums up our state of knowledge about the patient's condition: the i-th trinary digit of $s$ encode information about the symptom whose identifier is i. Let $\mathbb{A}$ be the action space:
$\mathbb{A}=\big\{a_1,...,a_{220}\big\}$. An action is a symptom that we suggest to the obstetrician to look for.

Our environment dynamic is clearly Markovian in the sense that: $P[s_{t+1}\mid a_t,s_t,a_{t-1},s_{t-1},...a_0,s_{0}]=P[s_{t+1}\mid a_t,s_t]$
where $s_t$ (respectively $a_t$) are the state (resp. the action) visited (resp. taken) at time $t$.

We aim to learn a diagnostic policy that associates each state of knowledge (list of presence/absence of symptoms) with an action to take (a symptom to check):

\begin{equation}
\label{policy}
\pi:\mathbb{S}\rightarrow \mathbb{A}.
\end{equation}

What should be a good diagnostic policy? Many medical applications consider a trade-off between the cost of performing more medical tests (measuring it in time or money) and the cost of a mis-diagnosis \cite{DBLP:journals/corr/abs-1109-2127}, \cite{Tang2016InquireAD}, \cite{Kao2018ContextAwareSC}.

However in our case the cost of performing more medical tests (i.e to check more possible symptoms) is negligible 
against the potential cost of a mis-diagnosis. In theory, the obstetrician have to check all possible symptoms to ensure the fetus does not present any disease. 
Therefore we will not take the risk of a mis-diagnosis by trying to ask fewer questions. However if the physician observes a sufficient amount of symptoms he can stop the ultrasound examination and perform additional tests, like an amniocentesis, to confirm his hypotheses. 

This is why we can label some states as terminal: they satisfy the condition that the entropy of the random variable disease is so low that we have no doubt on the diagnosis. In this setting, our goal is to minimize the average number of inquiries before reaching a terminal state: 
\begin{equation}
\label{myeq2}
\pi^*=\argmin_{\pi}\hspace{0.1cm} E_{\mathcal{P}}\big[ I \big| s_0, \pi\big],
\end{equation}
where $s_0=(2,...,2)$ is the initial state, $\mathcal{P}$ the law of the environment currently used, $\pi$ the diagnostic policy, and $I$ is the random number of inquiries before reaching a terminal states, i.e:



$$I=\inf \{t \mid H(D\mid S_t)\leq \epsilon \} $$
where $H(D\mid S_t)=\sum_{s_t} P[S_t=s_t] H(D\mid S_t=s_t)=-\sum_{s_t} P[S_t=s_t]\sum_d P[D=d\mid s_t]\log P[D=d\mid s_t]$ is the entropy of the random variable disease $D$ given what we know at time $t$: $S_t$. We should think $s_t$ as a realization of $S_t$, this is nothing more than the state we reached for one examination on a given patient while $S_t$ is the associated random variable. For a given start state $s_0$ and a policy $\pi$ there are many possible states that we can reach since the answers are stochastic. Note that we are not ensured that for all $t$ we had $H(D\mid s_{t+1})\leq H(D\mid s_t)$. Nevertheless this inequality holds when taking the average $H(D\mid S_{t+1})\leq H(D\mid S_t)$, see theorem 2.6.5 of \cite{Cover:2006:EIT:1146355}, "information can't hurt".  In summary, when we consider that entropy is sufficiently low and that we can stop and propose a diagnosis, we know that on average, the uncertainty about the patient's disease would not have increased if we had continued checking symptoms.


Setting a reward function as follow: $r_t:=r(s_t,a_t)=-1$, $\forall s_t, a_t$ we can rewrite \eqref{myeq2} in the classical form of an episodic reinforcement learning problem \cite{Sutton:2018:IRL:551283}:
\begin{equation}
\label{myeq1}
\pi^*=\argmax_{\pi}\hspace{0.1cm}E_{\mathcal{P}}\left[\sum_{t=0}^{I} r_t \Big|s_0,\pi \right].
\end{equation}

In the RL community such a reward design is called action-penalty representation, since the agent is penalized for every action that it executes \cite{Barto1995LearningTA}.

\subsubsection{Related works.}

There are numerous relevant expert system for the diagnostic of rare diseases (in particular in obstetric) such as for example Orphamizer see \cite{KOHLER2009457} and \cite{Khler2017TheHP}. Most of these expert systems use a list of observed symptoms as input and output a corresponding list of possible diseases. Nevertheless we think that an algorithm operable during the medical examination would be more useful than an expert system designed for retrospective use. This is why we aim to propose at each stage the most interesting symptom to check. 

Very few recent works try to address this issue. We reference \cite{DBLP:journals/corr/abs-1109-2127} which proposed an A* algorithm searching for shortest path in a graph but this kind of algorithm cannot cope with our high-dimensional problem. 

Note that in a certain sense our problem can be likened to a decision tree optimization task where the features are the symptoms and the disease is the target. Indeed a policy on a MDP is a generalization of a tree, a policy being less rigid in that sense that it can still propose the next feature to check when the physician made a different choice to the one we proposed. Classic decision tree algorithms, see \cite{Breiman1984ClassificationAR} or \cite{Quinlan:1986:IDT:637962.637969}, rely on optimizing an impurity function (the entropy or Gini index of the target random variable) in a greedy way and are therefore subject to the well-known horizon effect \cite{Berliner:1973:NCM:1624775.1624786}. Then recent works looking for global optimization procedure of decision trees such as \cite{Bertsimas:2017:OCT:3123655.3123731} can be seen as relevant. However, once again, these algorithms using MIO (Mixed Integer Optimization) solvers cannot cope with our high-dimensional problem. Indeed the complexity of such algorithms is $n\times 2^D$ where $n$ is the number of data and $D$ the maximal depth of the tree. Nevertheless in our case we can not restrict that easily the maximum allowed tree depth since in the worst case, the physician will not observe any symptom and will then have to check them all.    

More recent works of \cite{Tang2016InquireAD} and \cite{Kao2018ContextAwareSC} focus on this problem of building a symptom checker using reinforcement learning algorithms.
Nevertheless our approach is fairly different to these previous works, both in our way to formulate the objective (and then in our reward design) than in the solutions that we propose (our ways to break the dimension). 

They formulated their optimization problem as a trade-off between asking less questions and making the right diagnosis while we formulate it as the task of reaching as quick as possible, on average, a pre-determined high degree of certainty about the patient disease. In practice, in our case, the only parameter $\epsilon$ to be tuned is the degree of certainty we want at the end of the examination: we should stop when the entropy of the disease falls below this threshold. The fewer the $\epsilon$ the more symptoms our algorithm will need before considering that the game ends. 

\cite{Tang2016InquireAD} makes use of a discounted factor $\gamma\in [0,1]$ in their reward signal design. They design the reward associated to each question to be zero until possessing a diagnosis (which is an additional possible action) where the reward is equal to $\gamma^q$ (if the guess was correct, $0$ otherwise), $q$ being the number of questions that have been inquired before possessing the diagnosis. In this context $\gamma$ makes the compromise between asking fewer question and making the right diagnosis. The smaller $\gamma$, the more likely the algorithm is to make a wrong diagnosis by trying to ask fewer questions. 

Note that \cite{Tang2016InquireAD} has to perform its learning algorithm while trying several differents values of $\gamma$. On the contrary we can determine which value of $\epsilon$ we should take before launching any learning algorithms. We can indeed interact with the physician, in a first step, presenting him a sample of states where our algorithm would possesses a diagnostic. If the physician considers that the algorithm stops too early we should decrease $\epsilon$, otherwise we should increase $\epsilon$. This is an advantage since the main bottleneck in terms of computing time is the learning phase. 


\subsubsection{High-dimensional issues.}

Our full model is of very high dimension (220) and thus a classical tabular approach is impossible. According to our experiments, a classical Deep-Q learning is also not numerically tractable. In order to break the dimension, we capitalize first on the fact that the physicians use our algorithm mainly after seeing a first symptom. In such case, we make the assumption that this initial symptom is typical. It might be possible to have a disease which also presents a non-typical symptom but this happens with a very low probability, sufficiently negligible for the clinicians. Anyway, in this case, we would end up with a high entropy and no disease identification. This leads to switch to another strategy. With such an assumption the dimension drops significantly since we now only consider diseases for which this initial symptom is typical, the only relevant symptoms are the one which are typical of these remaining diseases.

Therefore we created $220$ tasks $\mathcal{T}_i$ to solve, $\forall i$, $s_{(i)}=(2,...,2,1,2,...,2)$: 
\begin{equation}
\label{subproblem}
\pi_{(i)}^*=\argmax_{\pi} E_{\mathcal{P}}\left[\sum_{t=0}^{I} r_t \mid s_{(i)}, \pi \right].\tag{$\mathcal{T}_i$}
\end{equation}

The different sub-problems dimensions are displayed in figure \ref{TaskDim}. Fragmenting that much our problem as the advantage of giving us a very good optimized policy on several part of our decision tree that would have been under-optimized otherwise (because these parts of the tree are not often visited). Of course, optimizing the parts of the tree that are not often visited is not very useful to reduce our overall loss function, but it is important to provide, in all cases, a reasonable proposal to the physician if we want him to have confidence in us. This approach will force us to choose a learning algorithm which can handle different subproblems without needing to tune too many hyper-parameters.      

To cope with these high dimensional issues, \cite{Tang2016InquireAD} proposed in their first paper to learn a different policy for each of the $11$ anatomical parts they previously built. As they recognized in their second paper \cite{Kao2018ContextAwareSC} this approach is problematic. Indeed a symptom may be related to several different anatomical parts. How to choose which model to use when observing an initial symptom? In their first paper, when a patient give an initial symptom, they choose the model with the best accuracy on their training set and follow this policy until the end of the process. Nevertheless, as they write in \cite{Kao2018ContextAwareSC}, it is possible that the target disease does not belong to the disease set of the chosen anatomical part. This is why they proposed to learn an other policy, called master model, which choose at each step the most promising model (among the $11$ anatomical parts) to use.  

\subsection{Two different approaches to solve our problem.}

In reinforcement learning, there exist several ways to solve a problem like \eqref{subproblem}. If the dimension is small enough it is possible to find the optimal solution explicitly using a dynamic programming algorithm \cite{Sutton:2018:IRL:551283}, for example using the value iteration algorithm. If the number of states is too high we have to parameterize the policy (policy-based approach) or to parameterized the Q-values (value-based approach). We have investigated both approaches to solve our problem.

\subsubsection{A policy-based approach with hand-crafted features as baseline:}\label{policybased}

In our application it can be interesting to propose several symptoms to check at the user, each with its corresponding score (interest to check it), instead of a single one. Indeed physicians might be reluctant to use a decision support tool which do not let them a part of freedom in their choice. This is why we consider an energy-based formulation, a popular choice as in \cite{pmlr-v24-heess12a}: 
\begin{equation*}
    \pi_{\theta}(s,a)=e^{\theta^T\phi(s,a)}\big/\sum_{b}e^{\theta^T \phi(s,b)}
\end{equation*} 
where $\pi_{\theta}(s,a)$ is the probability to take action $a$ in state $s$, $\phi(s,a)$ is a feature vector: a set of measures linked with the interest of taking action $a$ when we are in state $s$. To be more precise: $\phi(s,a)=\left(H(D\mid s)-E[H(D\mid s,a)],P[S_a\mid s],\mathds{1}_{A_a\in S_{max}(s)}\right)$ where $S_{max}(s)$ is the set of typical symptoms of the most likely disease at state $s$ and $H(D\mid s)$ is the entropy of the random variable disease at state $s$. In words $\phi(s,a)$ summarizes three reasonable way to "play" our game.

\begin{itemize}
\item Ask the question that minimizes the expected entropy of the disease random variable. This is exactly the \cite{Breiman1984ClassificationAR} way to play.
\item Ask the question where the probability of a positive answer is maximum. It is specific to our game where positive answers are much more informative than negative answers (it would not be the case in a classic $20$ questions game).
\item Inquire symptoms related to the currently most plausible disease. 
\end{itemize}


These features have been identified as relevant measures in rare disease research by physicians we are working with. They represent different way to think and dilemmas faced during medical examination: when I observed a symptom should I think about symptoms usually observed jointly or should I think about the most plausible disease and look for the corresponding symptoms? 
 
Note that this parameterized function $\pi_{\theta}$ is nothing more than a neural network without hidden layer designed with hand-crafted features. When properly optimized this policy outperforms, by construction, classical decision tree algorithm \cite{Breiman1984ClassificationAR}.

Our aim is to learn good parameters $\theta$ for each of our $220$ subproblems: $\theta_{(i)}^*=\argmin_{\theta} L_i(\theta):=E_{\pi_{\theta}}[ I \mid s_{(i)}].$
This kind of optimization problem, has been well studied by the reinforcement learning community, see \cite{Konda1999ActorcriticA} or \cite{Sutton1999PolicyGM} for the general analysis and \cite{pmlr-v24-heess12a} for the energy-based particular case. We trained our policy using a REINFORCE algorithm \cite{Williams1992SimpleSG}, since we have broken the dimension and that the number of parameters to learn is limited, this algorithm is perfectly suitable and exhibits similar performances to that of an Actor-Critic algorithm. 

\subsubsection{A value-based approach:}

\paragraph{Training Deep Neural Networks\\} We recall that the Q-values are defined as $Q_{\pi}(s,a)=E\big[\sum_{t'=t}^{I} r_{t'}\mid s_t=s,a_t=a,\pi\big]$, namely this is the expecting amount of reward when starting from state $s$, taking action $a$ and then following the policy $\pi$. The optimal Q-values, are defined as $Q^*(s, a)=\max_{\pi} Q_{\pi}(s, a)$ and satisfy the following Bellman equation: $Q^*(s,a)=E_{s'\sim\mathcal{P}}[r+ \max_{a'} Q^*(s',a')]$.   

The optimal policy $\pi^*$, is directly derived from $Q^*$: $\pi^*(s)=\argmax_{a}Q^*(s,a)$. Therefore we "only" need to evaluate $Q^*(s,a)$, $\forall s, a$. This can be done by a value-iteration algorithm which uses the Bellman equation as an iterative update: $Q_{i+1}(s, a) =E [r + \max_{a'}  Q_i(s',a')|s,a]$. It is known, see \cite{Sutton:2018:IRL:551283}, that $Q_i\rightarrow Q^*$ when $i\rightarrow\infty$.

As the dimension of the problem is too high to store/evaluate all the Q-values, we parameterized it by a neural network: $Q(s,a)\approx Q_w(s,a)$. 

The famous Deep Q-Network (DQN) algorithm proposed by \cite{Mnih2013PlayingAW} made possible the use of neural networks to parameterize the $Q$-values (then called $Q$-network) in the value iteration algorithm with function approximation. The $Q$-network, at iteration $i$, is trained by minimizing the loss function 
$L_i(w_i)=E_{s,a}\big[\big(y_i-Q_{w_i}(s,a)\big)^2\big]$ where $y_i = E_{s'\sim\mathcal{P} } [r+\max_{a'} Q_{w_{i-1}}(s',a')|s, a]$ is the target. This can be done by a standard back-propagation algorithm. In practice to successfully combine deep learning with reinforcement learning, the main idea is to use experience replay to break correlation between data: build a batch of experiences (transitions $s$, $a$, $r$, $s'$) from which one samples afterwards. Another trick is to freeze the target network during some iterations to overcome instability while learning.  

By doing so acting and learning are dissociated, the policy used to act (called behavior policy) is different from the one learned from the transitions sampled in replay memory (the target policy). In RL, this type of algorithm are called off-policy methods. It is a desirable propriety for a RL algorithm to be off-policy as the behavior policy will be designed to enforce exploration. 

Figure \ref{architectureDQN} shows the general simplified scheme of the algorithms used in deep reinforcement learning: an agent interact with its environment and collects data (transitions $s_t$, $a_t$, $r_t$, $s_{t+1}$) which are incorporated to the replay memory from which we sample to form the target policy. Periodically the behavior policy is updated with the current learned policy. In our case we update the behavior policy as soon as we made a gradient ascent step.

\usetikzlibrary{arrows,positioning}
\tikzset{
    >=stealth',
    punkt/.style={
           rectangle,
           rounded corners,
           draw=black, very thick,
           text width=6.5em,
           minimum height=2em,
           text centered},
    pil/.style={
           ->,
           thick,
           shorten <=2pt,
           shorten >=2pt,}
}

 
 
 


\begin{figure}[t]
\centering
\begin{tikzpicture}[node distance=1.2cm, auto,]
 \draw (-3,0) node[punkt] (Target) {\textbf{Target Policy and Loss Function}}; 
 \draw (0,2) node[punkt](Frozen){\textbf{Frozen network}}; 
 \draw (4,4) node[punkt](New){\textbf{New Data}};
 \draw (0,4) node[punkt](Behavior){\textbf{Behavior Policy}}; 
 \draw (-4,4) node[punkt](Current){\textbf{Current Network}}; 
 \draw (3,0) node[punkt](Replay){\textbf{Replay Memory}}; 

\draw[->,>=latex] (New) 
edge[pil,out=0,in=0] node {Store} (Replay);

\draw[->,>=latex] (Replay) 
edge[pil,out=180,in=0] (Target);

\draw[->,>=latex] (Target) 
edge[pil,out=180,in=180]  node[pos=0.3] {gradient ascent step}(Current);

\draw[->,>=latex] (Behavior) 
edge[pil,out=0,in=180] node {Play} (New);

\draw[->,>=latex] (Current) 
edge[pil,out=0,in=180] node {Update}(Behavior);

\draw[->,>=latex] (Frozen) 
edge[pil,out=270,in=10] node {}(Target);

\draw[->,>=latex] (Current) 
edge[pil,out=270,in=180] node[pos=0.3] {Update each C iterations}(Frozen);
\end{tikzpicture}

\vspace{1em}
\caption{General schema of Deep Reinforcement Learning algorithms}
\label{architectureDQN}
\end{figure}
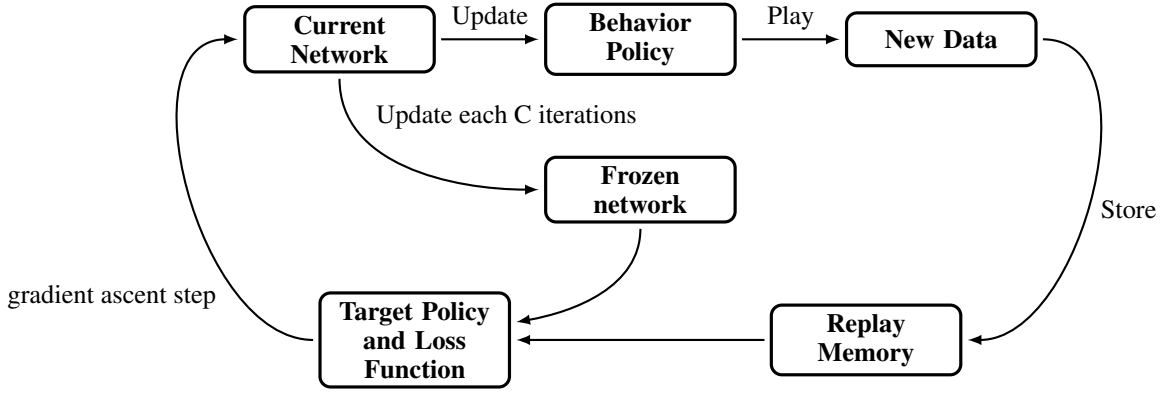



\paragraph{Some remarks on the behavior policy\\}
Our behavior policy is an $\epsilon$-greedy version of the current learned policy in order to enforce exploration. We use this policy to simulate games, or more precisely transitions $(s_t,a_t,s_{t+1},r_t)$. 

For this purpose, we need a model of the environment, a transition model which told us the probability to reach a state $s_{t+1}$ when taking action $a_t$ in $s_t$. Our environment model is composed of the symptoms combination distribution of each disease (see section \ref{modellearning}). Namely we store $(P[S_1,...,S_{K_D}\mid D])_{S_1,...,S_{K_D}}$ the probability of all the possible combinations of typical symptoms given the disease, $K_D$ is the number of typical symptoms of $D$. We add the assumption that a patient can also presents non-typical symptoms but with small probability and independently of the others symptoms (see section \ref{databasedefault}).

Then to simulate transitions we need to determine for each disease which are the symptoms of the current list which are typical and which are not. It will allow us to find the right combination we have to extract from $(P[S_1,...,S_{K_D}\mid D])_{S_1,...,S_{K_D}}$. This computation is not that cheap especially when we add ontological considerations (see section \ref{ontology}). We can speed it when we play games from the start $s_0$ to a terminal state $s_{I}$: we remember which symptoms are typical for each disease and thus only have to determine if the last symptom is typical or not.    

Note that, at each stage of a game, we have to compute the probability of the symptoms combination given each disease so as to determine whether we should stop or not. An other important observation is that trying to compute directly $p(s_{I}\mid s_0)$ is as costly as playing an entire game incrementally (as previously described) from $s_0$ to $s_{I}$. 

These two observations should convince the reader that an asynchronous learning approach, as in \cite{Mnih2016AsynchronousMF}, would not be suitable for our problem. From a computational perspective, it is reasonable to play games from the start to a terminal state.   

\paragraph{The update target: Temporal-difference and Monte-Carlo algorithm\\}

A remaining question concerns the definition of the update target, should we use Monte Carlo returns or bootstrap with an existing Q-function ?   

We recall (following \cite{Sutton:2018:IRL:551283}) that an algorithm is a bootstraping method if it bases its update in part on an existing estimate. This is the case of the Temporal-Difference (TD) algorithm defined as:
\begin{equation*}
Q_{k+1}(s,a) \leftarrow (1-\alpha) \underbrace{Q_{k}(s,a)}_{\text{old estimate}}+\alpha \underbrace{\left(r(s,a,s') +\max_{a'} Q_{k}(s',a')\right)}_{update}  
\end{equation*}
where we sampled $s,a,s'$ using the current policy (in a $\epsilon$-greedy way in order to enforce exploration) and the environment model $\mathcal{P}$. $Q_{k}$ is the estimate at iteration $k$, $\alpha$ the learning rate. On the contrary a Monte-Carlo method does not bootstrap:
\begin{equation*}
Q_{k+1}(s,a) \leftarrow (1-\alpha)Q_{k}(s,a)+\alpha G
\end{equation*}
where $G$ is the reward we received from a simulated game.

It is not clear at first sight whether we should use a TD method or a MC method to compute the target $y_i$. This question is the subject of a recent work \cite{amiranashvili2018td} which show that MC approaches can be a viable alternative to TD in the modern reinforcement learning era. Usually TD method is seen as a better alternative than MC method which is often  discarded because of the high variance of the return.  

Nevertheless our case study is specific: we face a finite-horizon task with a final reward: the reward signal is not very informative before reaching a terminal state. In addition, for the subproblems of intermediate dimensions, we are ensured that games will not last too much time and then that there is a small variance in the return of the Monte-Carlo episodes. 

We implemented both solutions referred as DQN-TD and DQN-MC. At each step of DQN-MC, we sample, following the behavior policy, $100$ games starting from the initial state $s_{(i)}$ and stopping when they reach a terminal state. All the transitions $s$, $a$, $s'$ of all these games are annotated with the reward they received (the number of questions that have been necessary to reach a terminal state during the game concerned) and incorporated in the replay memory. We then sample transitions from this replay memory (one twentieth) and perform  a gradient ascent step with a back-propagation algorithm (we used the Keras library \cite{chollet2015keras}).    

Concerning the DQN algorithm with TD method, we kept the main features of DQN-MC in order to facilitate their comparison. We play $100$ games, still with the behavior policy, and all the transitions $s$, $a$, $s'$ of all these games receive a $-1$ reward when $s'$ is not terminal, $0$ otherwise. The learning rate is initialized with a lower value than in the DQN-MC algorithm but it is decreased in exactly the same way in both cases: divided by two each $300$ iterations. Another difference is the frozen network we use as target in DQN-TD which is not needed in DQN-MC. We update the frozen network each $2$ iterations (we have also tried to update it less frequently but have not observed any major differences with the results presented here). 

We compared these two algorithms, DQN-MC and DQN-TD, on severals of our sub-tasks (see figures \ref{248} and \ref{1629}). We did not observed much difference on small and intermediate sub-problems: both algorithms converge at the same speed towards solutions of the same quality. Nevertheless DQN-TD appears much more sensitive to the learning rate. Indeed as it can be seen in figure \ref{248}, DQN-TD converge on this problem, where it remains $29$ relevant symptoms to check and $8$ possible diseases, when the learning rate is initialized at $0.0001$. Nevertheless if the learning rate is chosen a little bit higher, at $0.001$, DQN-TD diverge. On the contrary, DQN-MC converge when the learning rate is initialized to $0.001$ and also when initialized to $0.01$ even if the returns of the algorithm are less stable in this latter case. These observations have to be combined with the one of figure \ref{1629} where it remains $104$ relevant symptoms to check and $18$ possible diseases. We can see that in this case DQN-TD with an initial learning rate of $0.0001$ diverge. Reducing the learning rate to $0.00001$ does not change this fact. On the contrary we do not need to reduce the initial learning rate of DQN-MC (we take it equal to $0.001$) to make it converge to a good solution. Since we have to train as many neural networks as the number of sub-tasks, we need a robust algorithm able to deal with different task complexity without changing all the hyper-parameters.      

This is why we chose to use DQN-MC instead of DQN-TD. It is, indeed, a well-known issue sometimes referred as "deadly triad" \cite{Sutton:2018:IRL:551283} that combining function approximation, off-policy learning and bootstrap to compute the target (what the DQN-TD algorithm does) is not safe. We show that DQN-MC performs well on small and intermediate sub-tasks of our problem. The higher dimensional tasks are harder to solve because the games are expected to last longer which is a challenge both in term of computing time that in terms of learning stability (higher variance of the return). To scale up on such problems, we break down the state space into a partition and leverage already solved sub-tasks as bootstrapping methods. 

\begin{algorithm}
\caption{DQN-MC with Bootstrapping on already solved sub-tasks.}
\begin{algorithmic} 
\STATE Start with low dimensional tasks.
\FOR{i such that the task $\mathcal{T}_i$ has not been yet optimized}
\IF{$|\mathbb{S}_i|\leq 30$}
\WHILE{the budget for the optimization of this task has not been reached}
\STATE Play $100$ games ($\epsilon$-greedy) from the start $s_{(i)}$ to a terminal state.
\STATE Integrate all the obtained transitions to the Replay-Memory
\STATE Throws part of the Replay-Memory away (the oldest transitions of the replay)
\STATE Sample $1/20$ of the Replay-Memory
\STATE Perform a gradient ascent step (backpropagation algorithm) on the sample
\ENDWHILE
\ENDIF
\ENDFOR
\STATE
\STATE Continue with higher-dimension tasks.
\WHILE{there are still tasks to be optimized}
\STATE Choose the easiest task to optimize: the one with the highest proportion
\STATE of already solved sub-tasks (weighted by their probability to be faced)
\WHILE{the budget for the optimization of this task has not been reached}
\STATE Play $100$ games ($\epsilon$-greedy) from the start $s_{(i)}$ to a terminal state (condition (j)) 
\STATE or to a state that was yet encountered in an already solved task (condition (jj))
\IF{we stopped a game because of condition (jj)}
\STATE Bootstrap i.e use the network of the sub-tasks to predict 
\STATE the average number of question to reach a terminal state
\ENDIF
\STATE Integrate all the obtained transitions to the Replay-Memory
\STATE Throws part of the Replay-Memory away (the oldest transitions of the replay)
\STATE Sample $1/20$ of the Replay-Memory
\STATE Perform a gradient ascent step (backpropagation algorithm)\ENDWHILE
\ENDWHILE
\end{algorithmic}
\label{DQNMCBoot}
\end{algorithm}

\paragraph{Solving higher dimension tasks by bootstrapping with already solved sub-tasks\\}

We denote $\mathbb{S}_i=(S_{i_1},...,S_{i_k})$ the set of symptoms related with the symptom $i$, i.e this is the set of symptoms which are still relevant to check after observing the presence of symptom $i$. When $|\mathbb{S}_i|$ is small enough (say $|\mathbb{S}_i|<11$), we can learn the optimal policy $\pi^*$ by a simple Q-learning lookup table algorithm, see \cite{Sutton:2018:IRL:551283}.

Considering intermediate dimension problems (say $11<|\mathbb{S}_i|<31$)  we can use the DQN-MC algorithm which performs pretty well on these problems (see experiences in section \ref{DQN-MC}). 
For high-dimensional problems ($|\mathbb{S}_i|>30$) using directly the DQN algorithm would be time-consuming. An easy way to accelerate the learning phase of these big networks is to make use of the smaller networks previously trained. Indeed if $S_i$ is a symptom for which $|\mathbb{S}_i|$ is high, there must have some $S_j\in\mathbb{S}_i$ such as $|\mathbb{S}_j|$ is small enough and therefore such as the $Q$-values of $\pi_{(j)}^*$ have been yet computed or at least approached.
Put in another way, when we try to learn the optimal $Q$-network of a given problem, we yet know, for some inputs, the $Q$-values that should output a quasi-optimal $Q$-network.      

There are several ways to take advantage of these already optimized subtasks to optimize networks on larger tasks. A first idea would be to incorporate to the replay-memory of the larger task, the replay-memories of the already solved sub-tasks by having previously properly resized the states. Remind that at each iteration, i.e each gradient ascent step, we sample transitions from the replay-memory ($s$, $a$, $s'$ and the reward received at the end of the game $R$) to form the target and train set used to perform the back-propagation algorithm step. We can add to these sets some immovable transitions, the one we already know (because they appear in sub-problems already solved).     

However, by doing so we will face several issues. First, when we train our neural network using the replay memory constituted by playing on the concerned task, we are ensured that the transitions that populate our replay-memory will be present in a proportion equivalent to their probability of being encountered in the task. On the contrary, when we add some immovable transitions from already solved sub-tasks to our replay memory, we might over-optimize our network on these sub-tasks. Put it another way, the network will be over-optimized on parts of the decision tree which are not that frequently faced in practice.   

Secondly, although the length of the episodes will have been reduced since using the subtasks replay-memories allows us to learn more quickly how to play at the end of the games, it will still be time consuming to play from the beginning until the end of the episodes for tasks of high dimension. The length of the episodes will also be an issue considering the variance of the MC returns.   

Therefore a second idea would be to learn a policy on the higher dimension task by bootstrapping on already solved subtasks. Namely we play games starting from the initial state $s^{(i)}$ and bootstrap when reaching a state that belongs to a state set of the partition where there yet exist an optimized network. In practice we have a function which is called each time we received a positive answer which checks if there already exists a network optimized for such a starting symptom. If this is indeed the case, the current game is stopped and the corresponding  optimized network is called to predict the average number of question to ask to reach a terminal state. The main lines of the whole procedure are summarized in the algorithm \ref{DQNMCBoot}.  

Note that in doing so, we do not optimize the network for the entire task. It will therefore be necessary to change the neural network used for the recommendation during the examination when we change the space of the partition. The advantage is that we will not need to use a more complex architecture for this higher dimension task.

Finally, note that we are learning the $Q$-networks one after the other and that there is therefore a more preferable order than others for optimizing these deep networks. We choose at each step to optimize the $Q$-networks which has the highest rate of sub-problems already solved (where each sub-tasks is weighted by its probability to be faced).

\paragraph{Some remarks on the complexity of a task\\}

When we described the several tasks $\mathcal{T}_i$, we focused mainly on the number of remaining relevant symptoms to check denoted $|\mathbb{S}_i|$. This is the most important parameter since it is the input length of our network and then determines the number of network parameters that we will have to optimize. 

Nevertheless there are more parameters which influence the complexity of a task. Let us mention the number of possible diseases and especially their probabilities. Indeed, if there are many possible symptoms to check and many possible diseases but there is a disease that is much more plausible than the others, then the task is not so difficult. An other feature that can influence a task complexity is the amount of symptoms which are typical of several of the possible diseases. 

Thus as it seems difficult to quantify the difficulty of a task we should avoid to judge the performance of our algorithms in an absolute way but should always compare them to more classical methods.  

Finally note that even what we call "task complexity" is not that easy to define. An idea would be to define the complexity of a task as the difference between the average number of question that have to ask a random policy and the average number of question that have to ask the optimal policy. 

\subsection{Numerical Results.}

For all the experiments involving neural networks, we used the same architecture detailed in table \ref{neuralarchitecture}. We first use an embedding layer since the inputs processed by our neural network should not be treated as numerical values. We then use two hidden layer with ReLu activation and a final layer with linear activation which outputs the Q-values of the possible actions. The $\epsilon$ parameter of our stopping criterion is set to $10^{-6}$ for all the experiments.

\begin{table}[ht]
\caption{Neural network architecture for task $\mathcal{T}_i$. $|\mathbb{S}_i|$ the number of remaining relevant symptoms to check.}
\centering 
\begin{tabular}{c c c c} 
\hline\hline 
Name & Type & Input Size & Output Size \\ [0.5ex] 
\hline 
L1 & Embedding Layer & $|\mathbb{S}_i|$ & $3\times|\mathbb{S}_i|$ \\ 
L2 & ReLu & $3\times|\mathbb{S}_i|$ & $2\times |\mathbb{S}_i|$ \\
L3 & ReLu & $2\times |\mathbb{S}_i|$ & $|\mathbb{S}_i|$ \\
L4 & Linear & $|\mathbb{S}_i|$ & $|\mathbb{S}_i|$ \\
\hline 
\end{tabular}
\label{neuralarchitecture} 
\end{table}

\subsubsection{Our baseline has quasi-optimal performances on small subproblems.}\label{baseline}
We can compare the performance of our policies optimized by a REINFORCE algorithm, with a classic decision tree algorithm \cite{Breiman1984ClassificationAR} and also with the true optimal policies when it was possible to compute the latter, i.e. when the dimension was small enough. 

\begin{figure}[t]
    \begin{minipage}[c]{.46\linewidth}
        \centering
        \includegraphics[width=6cm]{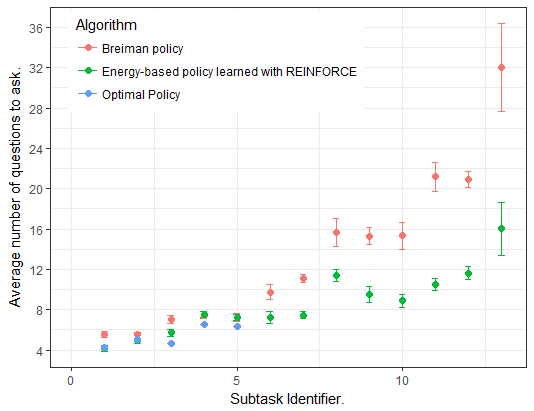}
        \caption{Average number of questions to ask on several subtasks.}
        \label{BreimanvsNous}
    \end{minipage}
    \hfill%
    \begin{minipage}[c]{.46\linewidth}
        \centering
        \includegraphics[width=6cm]{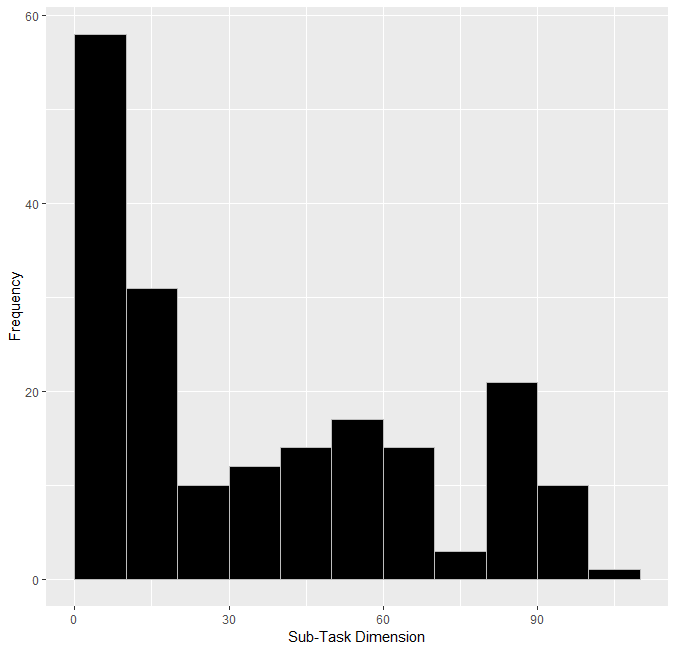}
        \caption{Dimensions of the different sub-tasks.}
        \label{TaskDim}
    \end{minipage}
\end{figure}

Results on some of our subtasks are presented in figure \ref{BreimanvsNous}. Our energy-based policy appears to clearly outperform a classic Breiman algorithm and all the more so as the dimension increases: the average number of questions to ask may be divided by two in some cases. On small subproblems where we have been able to compute the optimal policy by a dynamic programing algorithm, our energy-based policy appears to be very close to the optimal policy.    


\begin{figure}[t]
    \begin{minipage}[c]{.46\linewidth}
        \centering
        \includegraphics[width=6cm, height=4cm]{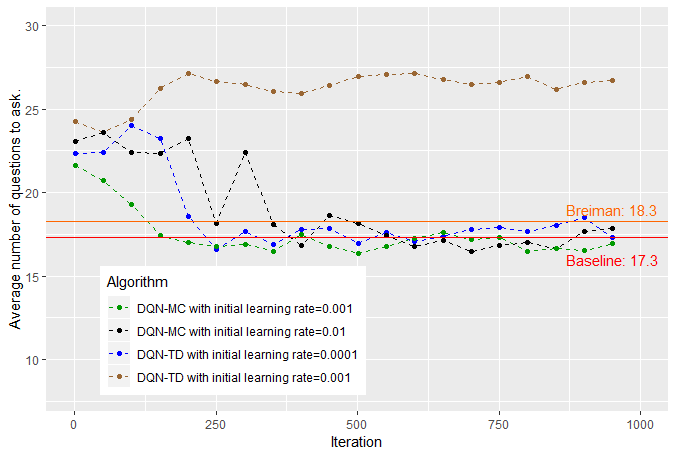}
        \caption{Comparison of DQN-TD and DQN-MC. Task dimension: $29$.}
        \label{248}
    \end{minipage}
    \hfill%
    \begin{minipage}[c]{.46\linewidth}
        \centering
        \includegraphics[width=6cm, height = 4cm]{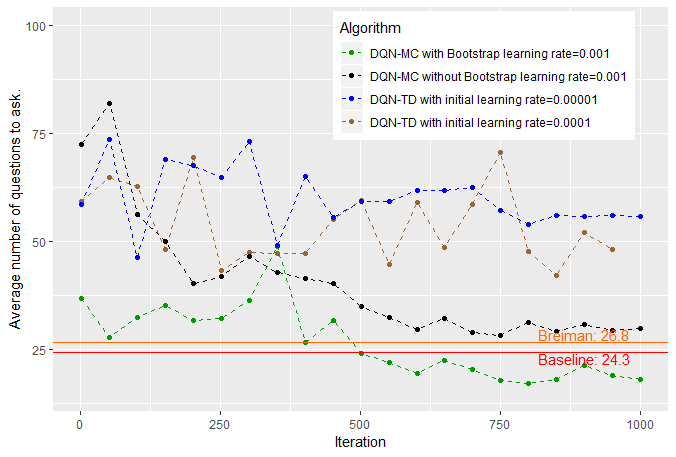}
        \caption{Comparison of DQN-TD, DQN-MC and DQN-MC-Bootstrap. Task dimension: $104$.}
        \label{1629}
    \end{minipage}
\end{figure}

\subsubsection{DQN-MC algorithm vs our baseline.}\label{DQN-MC}

We have performed a DQN-MC algorithm on our subtasks. We expect this algorithm to find a better path than the energy-based policy of section \ref{policybased} since a neural network has many more parameters and can therefore handle many more different situations than our baseline. Nevertheless to train such a high dimensional function instead of the three parameters of our baseline has a cost. How much iterations does need a DQN-MC to outperform our baseline? 

We recall here that an iteration of the DQN-MC algorithm consist in playing $100$ games that are added to the replay memory, then we sample one twentieth of this replay memory and perform a back-propagation algorithm. For comparison, our baseline has been trained with a REINFORCE algorithm, each iteration consist in playing one game and performing a gradient ascent step accordingly, we stop the training phase when reaching $1000$ iterations.      

\begin{figure}[t]
    \begin{minipage}[c]{.46\linewidth}
        \centering
        \includegraphics[width=6cm]{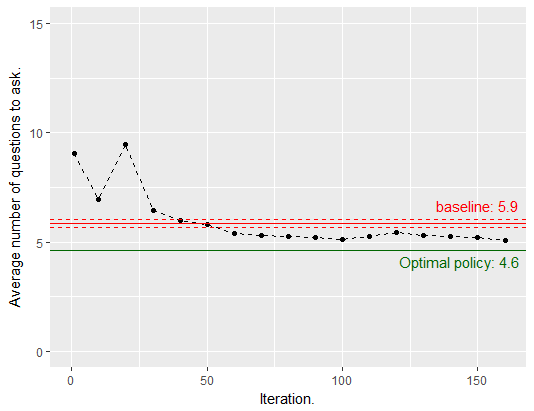}
        \caption{Evolution of the performance of the neural network during the training phase with DQN-MC. Task dimension: 10.}
        \label{268}
    \end{minipage}
    \hfill%
    \begin{minipage}[c]{.46\linewidth}
        \centering
        \includegraphics[width=6cm]{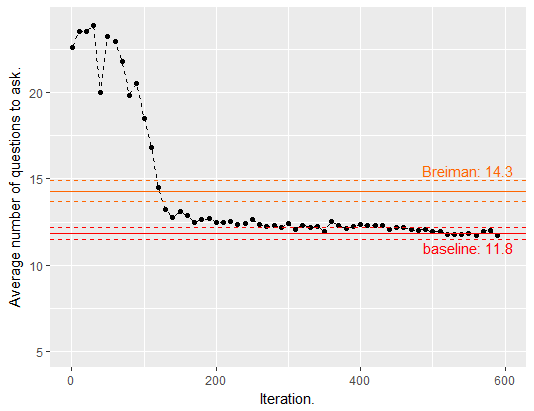}
        \caption{Evolution of the performance of the neural network during the training phase with DQN-MC. Task dimension: 26.}
        \label{45}
    \end{minipage}
\end{figure}

Figures \ref{268} and \ref{45} show, as expected, that the DQN-MC algorithm needs more simulations of games than our baseline. Indeed in these two sub-tasks, DQN-MC needed respectively $40$ and $200$ iterations to reach our baseline, so $40\times 100=4000$ and $200 \times 100=20000$ games instead of the $1000$ which trained our baseline. In figure \ref{268}, for a sub-task of dimension $10$, we can see that the DQN-MC algorithm needs a reasonable amounts of games to outperform our baseline. In that case, the DQN algorithm found a very good diagnostic policy but did not reach the optimal policy, it is probably stuck in a local extrema (although we do use an exploration parameter).

In figure \ref{45}, the DQN algorithm seems to converge toward the baseline. This might be due to the fact that, in these tasks of intermediate dimension (it remains $26$ relevant symptoms and $8$ diseases), our baseline is yet a good solution close to the optimal policy. Thus the DQN algorithm which is not ensured to converge to the optimal policy might get stuck in a local extrema at the level of the baseline. 

These experiments can be conducted in a laptop without use of GPU and should be then easily reproducible using our environment simulator or a similar one.

Finally, as one might expect considering the difference of needed iterations to converge between figure \ref{268} and figure \ref{45}, the idea of using previously resolved subtasks  will be important to deal with high-dimensional tasks.       

\subsubsection{Bootstraping on already solved sub-tasks helps (a lot) for high-dimensional tasks}

In these experiments, we compare the performance of a simple DQN-MC algorithm against a DQN-MC-Bootstrap on some of our tasks. We used the same neural network architecture for both algorithms (see table \ref{neuralarchitecture}). More broadly the two algorithms use exactly the same hyper-parameters, the only difference being the bootstrap trick of DQN-MC-Bootstrap. 

\begin{figure}[t]
        \centering
        \includegraphics[width=6cm]{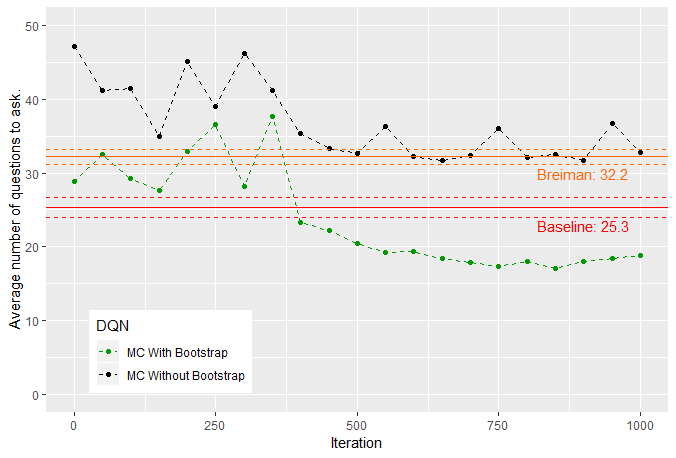}
        \caption{Evolution of the performance of the neural network during the training phase. Task dimension: $70$.}
        \label{1561}
\end{figure}

Figures \ref{1561} and \ref{1629} show the benefits of using the solved sub-tasks as bootstraping methods. In both cases a simple DQN-MC is unable to find a good solution while a DQN-MC-Bootstrap outperforms pretty quickly our baseline. Note that the neural network trained with DQN-MC-Bootstrap starts with a policy that is not that bad. It is appreciable as it reduces, since the beginning of the training phase, the length of the episodes and then the computing cost associated. 

For the experiment of figure \ref{1561} it remains $70$ relevant symptoms to check, $9$ possible diseases including the disease "other", and $20$ sub-tasks have been already solved. Finally the probabilities of presence of each of the subtasks initial symptom given the initial symptom of the main task were (0.01; 0.44; 0.01; 0.15; 0.15; 0.01; 0.03; 0.02; 0.11; 0.01; 0.26; 0.01; 0.03; 0.01; 0.15; 0.01; 0.15; 0.24; 0.16; 0.06).

For the experiment of figure \ref{1629} it remains $104$ relevant symptoms to check, $18$ possible diseases including the disease "other", and $103$ sub-tasks have been already solved. 

Finally we have been able to learn a good policy for the main task \eqref{myeq2} where it remains $220$ relevant symptoms to check, $82$ possible diseases including the disease "other"  and all the possible sub-tasks have been already solved. Our DQN-MC-Bootstrap algorithm starts with a good policy which only needs $45$ questions on average to reach a terminal state. Some training iterations allows it to improve until needing $40$ questions to reach a terminal state. On the contrary the experiment we made on a DQN-MC which tries to solve from scratch this task has to ask $117$ questions, on average, to reach a terminal state and does not improve significantly during the $1000$ iterations. We have evaluated also the performance of the Breiman policy on the global task, it needs $89$ questions on average to reach a terminal states (with a variance of $10$ questions).       

\subsubsection{A qualitative analysis for a low-dimensional sub-task}

\begin{figure}[t]
        \includegraphics[ width=14.5cm,height=5.5cm]{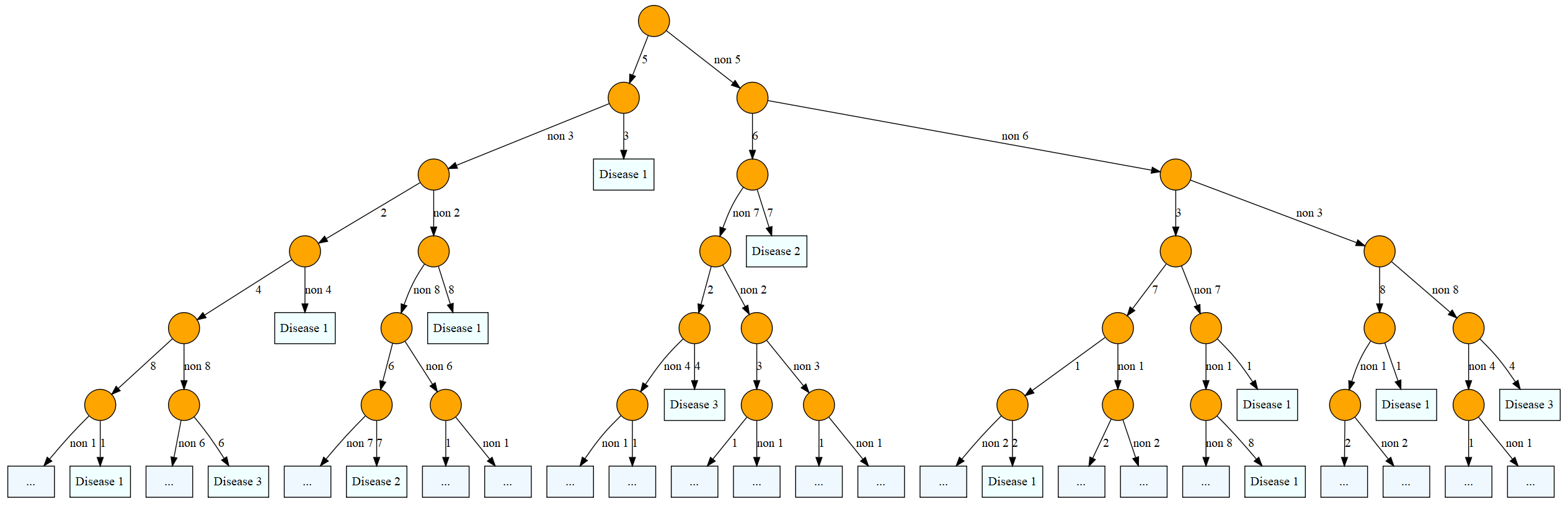}
        \caption{Optimal decision tree for a subtask with $8$ possible symptoms.}
        \label{decisiontree}
\end{figure}

We analyze here the policy obtained by using a look-up table value iteration algorithm on a small sub-task (it remains $8$ relevant symptoms to check) in order to illustrate some of the dilemmas a medical doctor can face during an examination. We start with the presence of symptom $9$. The three diseases which does have symptom $9$ in their list of typical symptoms are displayed in table \ref{disease-table}. We should think, for this one experiment only, that the symptoms are conditionally independent given the disease. An other important information is the prevalence of each disease, we have $P[D=d_1]=0.042$, $P[D=d_2]=0.0083$ and $P[D=d_3]=0.0083$. Finally there is no relation of ascendant/descendant between the $9$ symptoms considered in this example. The optimal strategy obtained induces a decision tree  which is displayed in the figure \ref{decisiontree}.   

\begin{table}[h]
  \caption{List of plausible diseases and corresponding list of related symptoms for the sub-task starting with presence of symptom $9$.}
  \label{disease-table}
  \centering
\begin{tabular}{cccccc}
\toprule
\multicolumn{2}{c}{Disease 1} & \multicolumn{2}{c}{Disease 2} & \multicolumn{2}{c}{Disease 3} \\
\cmidrule(lr){1-2} \cmidrule(lr){3-4} \cmidrule(lr){5-6}
Id Symptom & Probability & Id Symptom & Probability & Id Symptom & Probability\\
\midrule
 1 & 0.50 & 6 & 0.90 & 2 & 0.90\\
 2 & 0.55 & 7 & 0.50 & 4 & 0.90\\
 3 & 0.50 & 9 & 0.90 & 6 & 0.50 \\
 5 & 0.90 &  &  & 9 & 0.50 \\
 8 & 0.50 &  &  &  & \\
 9 & 0.50 &  &  &  & \\
\bottomrule
\end{tabular}
\end{table}

The first question is comprehensible, it ask about the most plausible symptom of the most plausible disease: the symptom $5$. If the answer was positive it continue with a symptom typical of the first disease which is not also typical of other diseases: the symptom $3$. The combination of the presence of this two symptoms is sufficient to diagnose the disease $1$. The rest of the tree is less obvious. For example when we get a "yes" for symptom $5$ and a "no" for symptom $3$, should we continue asking symptoms related to disease $1$ or should we switch to the symptoms typical of the other diseases ? The founded path chooses a symptom related to both disease $1$ and disease $3$ (symptom $2$), probably because it is then easy (and fast) to discard disease $3$ by asking a question about symptom $4$ (note that the disease $3$ has only $3$ related symptoms). 

An other interesting parts of the tree is when we received a negative answer to our first question about the symptom $5$. Then, the initial most plausible disease (the disease $1$) becomes less likely, but it is not clear if its probability decreases that much that we should check symptoms of other diseases or not. In this case the optimal strategy is to switch to symptoms of disease $3$ which has less typical symptoms and must be (in this part of the tree) more plausible than the disease $1$.   

We do not draw the entire decision tree for visibility reasons, we wrote "..." for the leaves where the obtained diagnostic strategy still propose to check more symptoms. 

\section{Learning a model of the environment.}
\label{modellearning}
\subsection{The need to learn a model}

As described in section \ref{planning} our agent will be trained while interacting with its environment. We focused until now on the planning task: optimizing the policy using observed transitions (initial state, action, reached state, reward)$=(s_t,a_t,s_{t+1},r_t)$. 
We should now detail how these transitions, these data, can be obtained. There exist several possible approaches in reinforcement learning:   
\begin{itemize}
\item \textbf{Model-based RL}: We first build a model of the environment in order to know how our environment will react to our actions. Then our agent is trained using experiences simulated from this model (planning task). 
\item \textbf{Model-free RL}: We do not try to infer the environment dynamic, we just train our agent using trial-and-error directly obtained by the interaction with the environment.  
\end{itemize}

\usetikzlibrary{arrows,positioning}
\tikzset{
    >=stealth',
    punkt/.style={
           rectangle,
           rounded corners,
           draw=black, very thick,
           text width=6.5em,
           minimum height=2em,
           text centered},
    pil/.style={
           ->,
           thick,
           shorten <=2pt,
           shorten >=2pt,}
}

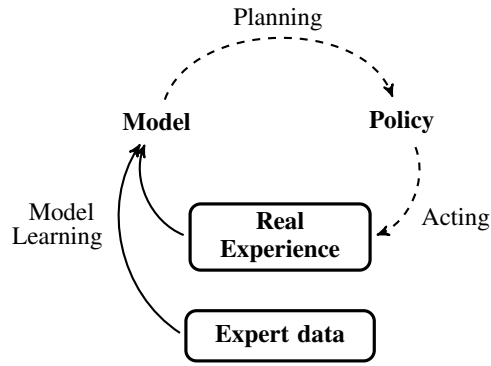
\begin{figure}[t]
\centering
\begin{tikzpicture}[node distance=0.95cm, scale=0.6,auto, every node/.style={scale=0.95}]
 \node[punkt] (market) {\textbf{Real\\ Experience}};
 \node[punkt, inner sep=5pt,below=0.5cm of market]
 (formidler) {\textbf{Expert data}};
 \node[above=of market] (dummy) {};
 \node[right=of dummy] (t) {\textbf{Policy}}
   edge[pil,bend left=45, dashed] node[auto]{Acting} (market.east); 
 \node[left=of dummy] (g) {\textbf{Model}}
   edge[pil, <- ,bend right=45]node[auto,xshift=-51pt,yshift=-13pt]{Model}(market.west)
   edge[pil, <-, bend right=45] node[auto,xshift=-48pt,yshift=-5pt]{Learning}(formidler.west)
   edge[pil,->, bend left=70, dashed] node[auto] {Planning} (t);
\end{tikzpicture}

\vspace{1em}
\caption{Our global architecture.}
\label{architecture}
\end{figure}

We can not adopt a model-free RL approach for obvious reasons. Beyond ethical considerations (one would use, at the beginning, an algorithm without any knowledge on real patients), a model-free architecture would need a very large amount of data/time to learn a good policy especially considering the diversity of situations it will face. This is a time that domain knowledge can save us.

Therefore we will need to learn a model of the environment. This model learning phase can often be avoided. For example in adversarial games a popular solution is to use self-play. This is the case in recent advances of computer Go \cite{Silver2017MasteringTG} which shows that it is possible to achieve a superhuman level in a challenging domain as Go without any domain knowledge, using only reinforcement learning with self-play. However, we are not in an adversarial game where we could learn from self play.
Another approach is to use expert demonstrations in order to estimate both rewards and environment dynamics \cite{pmlr-v51-herman16} or to learn directly a policy \cite{2013arXiv1307.3785T}. Expert demonstrations are often integrated as a supervised learning initialization step in the AI architecture as in \cite{Silver2016MasteringTG}. We do not have such expert demonstrations and in any case, since we are interested in rare diseases, we would need a very large amount of demonstrations to learn a good policy.  

Taking into account the uncertainty in the transition model has been tackled by model-based Bayesian reinforcement learning theory \cite{DBLP:journals/corr/GhavamzadehMPT16}. The main idea is to put a prior on the unknown transitions probabilities and update them when observing transitions in the real world. This approach is not suitable in our case since we do not have a prior on transition probabilities but rather on symptoms marginal distributions which is a less classical prior form (see section~\ref{data}).

Generally speaking our application area is specific by its lack of data, making the environment dynamic very uncertain. This prevent us from designing our architecture without any domain knowledge. 

We detail in the following section the model learning phase of our architecture (see figure~\ref{architecture}) where we integrate expert data to the data collected by the experience of the algorithm in order to build a sufficiently accurate model of the environment.

\subsection{Transition model learning: from marginal to joint distributions.} \label{learnmodel}

\subsubsection{Our approach: a trade-off between expert and observations.} \label{optimproblem}

Our problem is that we only know the symptom marginal distribution given the disease and not their joint distribution.  
We have $P[S_i\mid D]$ $\forall i,\forall D$ but need $P[S_{i_1},...,S_{i_{K_D}}\mid D]$ $\forall i_1,...,i_{K_D}, \forall D$.

We do not want to make the assumption of conditional independence  since we expect complex correlations between symptoms for a given disease. Note that the  assumption of conditional independence would make it possible to present a disease without having any of the symptoms related to this disease in the database (when there is no $S_i$ such that $P[S_i\mid D]=1$), which should be impossible. 
  






We emphasize that the knowledge of $P[S_i\mid D]$ does not give information regarding $P[S_{i_1},...,S_{i_{K_D}}\mid D]$ when conditional independence is not true. We can imagine two symptoms individually very plausible but who rarely occur together (or even never in the case of incompatible symptoms as for example microcephaly and macrocephaly). We chose to give values to  $P[S_{i_1},...,S_{i_{K_D}}\mid D]$ such as to maximize the entropy of the distribution $\left(P[S_{i_1},...,S_{i_{K_D}}\mid D]\right)_{i_1,...,i_{K_D}}$ under constraints given by the marginals. Indeed we have to add information but as little as possible on what we do not know. This approach is called maxent (maximum entropy) see \cite{JaynesInformationTA},  \cite{Cover:2006:EIT:1146355}, \cite{Berger:1996:MEA:234285.234289}. 

Our approach is Bayesian since it assumes knowing some properties of the distribution to be estimated (traditionally its mean, in our case its marginals) and looks for the maximal entropy distribution which verifies these constraints. Note that without any additional constraint, the distribution of maximum entropy with fixed marginal is the independent one. However we can add some information about the structure of the desired distribution as constraints in our optimization problem. We judge impossible to have a disease without having at least a certain amount of its associated symptoms: one, two or more depending on the disease. Indeed the disease we are interested in manifest themselves in combination of symptoms. 

Moreover our algorithm not only relies on expert data but also uses data collected from its own experience.
If we had enough data from direct experiments of the algorithm, we would not need expert data anymore. On the contrary without experimental data our model should rely entirely on expert data.  

Let's write \begin{equation}\label{vecteurinconnu}
\mathcal{P}=\left(\begin{matrix} P[S_1\mid D] \\ \vdots \\P[S_K \mid D] \\ P[\bar{S_1},...,\bar{S_K}\mid D]\\ \vdots \\ P[S_1,...,S_K\mid D] \end{matrix}\right)=:\left(\begin{matrix} P(1) \\ \vdots \\P(K)  \\ \pi_{1} \\ \vdots \\ \pi_{2^K} \end{matrix}\right)
\end{equation}
the vector we aim to estimate: the symptom distribution of a disease with $K$ associated symptoms and its marginals. We propose to estimate it with the following optimization problem:

\begin{align}
\mathcal{P}^{new}=&\argmax_{\mathcal{P}/\mathcal{P}\in \mathcal{C}} \hspace{0.1cm} L\left(x^{(1)},...,x^{(N)}\mid \mathcal{P}\right) \label{myeq3} \\  
&+\epsilon\left(H(\pi)
- \sum_i \lambda_i KL\Big(Be\left(P^{expert}(i)\right)||~Be\big(P(i)\big)\Big)\right)\nonumber\\
=:&\argmax_{\mathcal{P}/\mathcal{P}\in \mathcal{C}} \hspace{0.1cm} J(\mathcal{P}) \nonumber
\end{align}

where the constraint $\mathcal{P}\in \mathcal{C}$ just states the classical probability measure constraints: respect of marginals and sum equal to one, we also add the constraint to set to $0$ symptoms combinations considered impossible. We have three terms:  

\begin{itemize}
    \item First a log-likelihood term for experimental data: $L\left(x^{(1)},...,x^{(N)}\mid \mathcal{P}\right)$ where $x^{(i)}$ is the i-th combination of symptoms observed in real life. We aim at maximizing this quantity since we want our model to be coherent which what we observed. Symptoms combinations observed in real life should be considered a little bit more plausible. Note that the log-likelihood of independent observations $x=(x^{(1)},...,x^{(N)})$ under model $\mathcal{P}$ has a very simple form:    
$$L\left(x^{(1)},...,x^{(N)}\mid \mathcal{P}\right)=\sum_{i=1}^{2^K} N_i \log(\pi_i)$$
 where $N_j(x)=\displaystyle\sum_{k} \mathds{1}_{\{x^{(k)}=j\}}$ is the number of times we had observed the j-th symptom combination.
\end{itemize}

However we can not just maximize the likelihood since we do not expect to have sufficient amount of data to infer symptoms distributions. Note that in the worst case when a disease has $19$ possible symptoms there are $2^{19}\approx 500.000$ possible combinations of symptoms. It is far too much to infer the symptom distribution with a maximum likelihood approach, especially considering data scarcity.  
\begin{itemize}

\item This is why we add an entropy term, $H(\pi)$, in order not to consider impossible a symptom combination that has not been yet observed in real life.
\item The last term ensures that the marginals of our new distribution will not stray too far from our initial a priori given by expert data: $P^{expert}(i)$. We recall that $KL(P||Q)\geq 0$, $\forall P,Q$ and $KL(P||Q)=0\Leftrightarrow P=Q$  
Note that each marginal does not have the same coefficient $\lambda_i$ as we do not have the same confidence in all the expert data. In particular we can handle missing data, i.e when we do not know $P[S_i\mid D]$, by setting $\lambda_i=0$.
\end{itemize}

\subsubsection{Existence/uniqueness of a solution and numerical considerations.}
The function $J$ defined in equation \eqref{myeq3} we aim to optimize is $\mathcal{C}^{\infty}$ on the constraint space $\mathcal{C}$ which is a compact set \big(since $\mathcal{C}\subset [0,1]^{2^K+K}$\big), therefore $J$ admits a maximum in $\mathcal{C}$. As $J$ is concave (as a sum of concave functions) this maximum is unique and we can use the Kuhn-Tucker theorem which ensures us that maximizing our function under constraints can be achieved looking for the saddle-point of the Lagrangian.     
%
%
%
%
Deriving the Lagrangian and equating it to $0$, we obtain the marginals as function of Lagrangian parameters $\mu$. We write $\mu=(\mu_0,\mu_1,...,\mu_K)$ the Lagrangian parameters where $\mu_0$ states for the constraint $\sum_j \pi_j=1$ and each $\mu_k$ states for the marginal constraint respectively to $P(k)$. If $\lambda_j\neq 0$ we have:

$$P(j)=\left(1+\displaystyle\frac{1}{\left(\displaystyle\frac{P^{expert}(j)}{1-P^{expert}(j)}\right)\exp\left(\displaystyle\frac{\mu_{j}}{\epsilon \lambda_j}\right)}\right)^{-1};$$

Note that if $\epsilon \lambda_j \rightarrow +\infty$ we indeed recover $P(j)=P^{expert}(j)$.

Moreover if $N_j=0$ we have:

\begin{equation}
\label{solutionmaxent}
 \pi_j=\exp\left(-1-\displaystyle\frac{\mu_0}{\epsilon}-\displaystyle\sum_{k}\displaystyle\frac{\mu_k}{\epsilon}\mathds{1}_{\{S_k=1\}}\right);
\end{equation}

If $N_j\neq 0$ we can not obtain a closed form for $\pi_j$ as function of $\mu$ and we have to solve the following equation:
$$-\epsilon \left(\log(\pi_j)+1\right)+\mu_0+\sum_{j=1}^{K}\mu_j \mathds{1}_{\{S_j=1\}}+N_j \displaystyle\frac{1}{\pi_j}=0;$$
A dichotomy method will be suitable for this task. 

Readers familiar with maximum entropy theory should not be surprised by the form of equation~\eqref{solutionmaxent} . We recover a classical result, see for example \cite{Berger:1996:MEA:234285.234289}, the solution of maxent have a nice exponential form: a Gibbs distribution.  
  
We use an Uzawa algorithm to reach the saddle-point of the Lagrangian, see \cite{uzawa1958imc}. Since $J$ is a concave function we are ensured that the saddle-point we converge to by Uzawa iteration is the global maximum of $J$. 

\subsubsection{Heuristics for parameters choice.}

There are two kind of parameters to choose: $\epsilon$ and $\lambda_j$, $\forall j \in [1,K]$.

We could think that $\epsilon$ should decrease with $N$, but as we have chosen not to renormalize the log-likelihood we have  $L\big(x^{(1)},...,x^{(N)}\big)\rightarrow \infty$ when $N$ goes to infinity. A $\epsilon$ parameter independent of $N$ seems an easy calibration which provides good results, it should just be chosen large enough to regularize the log-likelihood when $N$ is small (see experiences in section~\ref{avecdata}).    

However $\epsilon$ should depend on the number of unknown parameters of the distribution to be estimated: $2^K$ \big($K$ the number of typical symptoms\big). Indeed a disease with $12$ typical symptoms (i.e. $2^{12}$ possible symptoms combination) will need far more data than a disease with $4$ typical symptoms. 

To calibrate $\epsilon$ as a function of $K$, we should look at how the three different terms of \eqref{myeq3} behaves with $K$. Roughly speaking the entropy term $H(\pi)$ is of the order of $K$ (the maximal values is $K\log(2)$ reached by the uniform distribution). The Kullback-Leibler penalization is linear in $K$ and appears scalable to the entropy (see section \ref{sansdata}).   

As it is usually done in log-likelihood regularization we expect the log-likelihood to be of order $2^K$: therefore, $\epsilon=c\times 2^K$
where $c$ is a non-negative constant to be determined, seems a reasonable calibration. In practice, our AI will never have a sufficient amount of data and the maxent regularization will allow us to cope with new situations. We will take this into account when choosing $c$.  

Concerning $\lambda_j$ parameters, the more confident we are in $P^{exp}(j)$ the higher is $\lambda_j$. We simply have to initialize $\lambda_j$ with sufficiently large values in order to prevent the condition on high entropy to change the marginals too much when $N$ is small as we will see in section~\ref{sansdata}. Of course we should not fall into the opposite excess by taking $\lambda_j$ too large which would have the consequence of staying on the experts' a priori even when the data tell us another reality. 

\subsubsection{Some previous works.}

Building a decision support tool in medicine has been an objective since the beginning of the computer age. Many of these early works proposed rules-based expert system but in the 80's an important part of the community investigated probabilistic reasoning based expert system \cite{Pearl1989ProbabilisticRI}. Probabilities and Bayesian methods were seen as a good way to handle uncertainty inherent to medical diagnosis. 

The conditional independence assumption of symptoms given the disease has been extensively discussed as it is of crucial interest in terms of computational tractability. Some researchers considered this assumption harmless \cite{Charniak1983TheBB} when others already proposed a maxent approach to face this issue \cite{Hunter:1985:URU:3023810.3023813}, \cite{DBLP:journals/corr/abs-1304-3423} or \cite{DBLP:journals/corr/abs-1304-1104}.

Nevertheless it seems that none of the works of that time has ever considered the experts vs observations trade-off we face. In the survey \cite{DBLP:journals/kbs/Jirousek90} it is clearly mentioned that these methods only handle input data of probabilistic form. Namely they assume to have an a priori on marginals but also on some of the possible probabilities combinations (in our case we would assume to have an a priori on $P[S_1,S_2\mid D]$ for example) and propose a maxent approach where these input data are treated as constraints in the optimization process. Once again this is not our case since we just have the marginals and some experimental data. This area of research was very active in the 80s and then gradually disappeared, probably due to computational intractability of the proposed algorithms.  

Estimating a joint distribution from marginals is an other very ancient problem, not necessarily related to AI, known in literature as cell probabilities estimation problem in contingency table with fixed marginals (the book \cite{Bishop75discretemultivariate} provide a good overview of this field). We can trace this problem back to the works of \cite{deming1940} which make use of known assumed marginals values and experimental samples to try to estimate the joint distribution. They proposed an iterative proportional fitting procedure (IPFP), a particularly popular algorithm, to solve this problem.

An important assumption of \cite{deming1940} is that each cell of the contingency tables received data. 
In \cite{Ireland1968ContingencyTW} the authors proved that the asymptotic estimator obtained by IPFP algorithm minimizes the Kullback-Leibler divergence with respect to the empirical distribution under the marginal constraints.  

However an IPFP algorithm would not be suitable for our problem for two main reasons: first we do not have absolute confidence in marginals given by experts and secondly we are interested in rare diseases so we do not expect to have a sufficient amount of data. As a matter of fact, many cells will not receive data, and it would be disastrous to assign $0$ to the corresponding symptom combination probability estimate in our application.       

We should mention also works that relate to our problem in applications of statistics to social sciences where researchers aim to build a synthetic population with marginals coming from several inconsistent sources \cite{Barthelemy2013SyntheticPG}. To be more precise they have data at an aggregated level (at the level of the country for example) an need disaggregated data (at the level of a household say). They also proposed a maxent approach but do not exactly face an expert/experience trade-off since they build their model without samples. Their algorithm is different essentially because they add constraints to obtain an integer solution, which we believe could be avoided. 

\subsection{Some experiments}

\subsubsection{Maxent with Kullback penalization on marginals a priori.}\label{sansdata}

Let us start by looking at what happens when we make a maxent with Kullback regularization on marginals, i.e we exclude likelihood for this synthetic experiment. Namely we are interested in a vector $\mathcal{P}^*$ defined as follows:
$$\mathcal{P}^{*}=\argmax_{\mathcal{P}/\mathcal{P}\in \mathcal{C}} \hspace{0.1cm}  H(\pi)
- \lambda \sum_{i=1}^{K} KL\Big(Be\left(P^{expert}(i)\right) ||~Be\big(P(i)\big)\Big);$$
We set $K=4$, with the following a priori on marginals  $\big(\frac{9}{10},\frac{8}{10},\frac{3}{10},\frac{2}{10}\big)=:\big(P^{expert}(1),P^{expert}(2),P^{expert}(3),P^{expert}(4)\big).$

\begin{figure}
   \begin{minipage}[c]{.475\linewidth}     \includegraphics[scale=0.675,width=6.75cm,height=6.75cm]{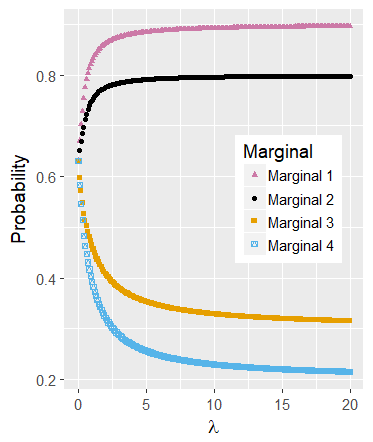}
   \caption{Evolution of marginals’ estimates in maxent with marginals regularization.}
   \label{maxentsansdonnees}
   \end{minipage} \hfill
   \begin{minipage}[c]{.475\linewidth}     \includegraphics[scale=0.65,width=6.50cm,height=6.50cm]{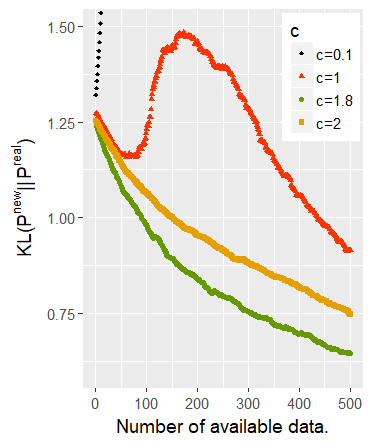}
      \caption{Evolution of the divergence between our
estimate and the real symptom combination distribution
as function of the amount of available data.}
      \label{performancedonnees}
   \end{minipage}
\end{figure}

For this experiment, we increase $\lambda$ and look at how it affects the marginals' estimates. We can see in figure \ref{maxentsansdonnees} that all the marginals' estimators start with value $0.63$ and then decrease or increase in a monotonous way toward their a priori. This is not surprising since maxent tends to disseminate weight on the entire distribution making parameters of Bernoulli marginals distribution closer to $0.5$. In our case we have marginals equal to $0.63$ since we enforce combinations with less than one symptom to have zero probability. This gives us an idea of how $\lambda$ should be initialized.

\subsubsection{Adding data.}\label{avecdata}

We simulated a symptom combination distribution $\mathcal{P}^{real}$ (with $K=8$ associated symptoms) using Poisson distribution of parameter $1$. The estimate $\mathcal{P}^{new}$ solution of \eqref{myeq3} given by our Uzawa algorithm has been sequentially updated using data sequentially simulated from $\mathcal{P}^{real}$. For the a priori on marginals, we used the real marginals with an additive Gaussian noise of zero mean and $1/4$ variance. The measure of interest is the Kullback-Leibler divergence between the real distribution and our estimate solution of~\eqref{myeq3}: $\text{KL}(\mathcal{P}^{new}||\mathcal{P}^{real})$ which we would like to minimize. We are interested in how the choice of $\epsilon=c\times 2^K$ affects our estimation of the real distribution. To cope with inherent randomness of this process, an average estimate of the Kullback-Leibler divergence was obtained over 50 repetitions of the same procedure (i.e we simulated $50$ Poisson distributions for each different values of $c$). 

In Figure~\ref{performancedonnees}, the red ($c=1$) and black ($c=0.1$) curves clearly show that giving too much weight to the data leads to over-weighing the symptoms combinations observed in real life and keeps us far from the real distribution: we do not sufficiently regularize with the entropy. On the contrary the green ($c=1.8$) and the orange ($c=2$) curves performs a good trade-off maxent/maximum likelihood. $c=2$ is a more cautious choice (we underweight experimental data) than $c=1.8$ and as a consequence the procedure converge less quickly to the real distribution.  

Note that an empirical estimate (solution of a maximum likelihood approach) or an IPFP algorithm would perform very poorly on this task. Indeed many symptoms combinations would be estimated to be $0$ when they should not, because of data scarcity: indeed we have $2^8=256$ variables and less than $500$ data. We have not plotted the Kullback-Leibler divergence of these estimates with respect to the real distribution since it is infinite. In contrast our approach appears robust to data scarcity, provided that we take care of the value of $\epsilon$.   

\subsection{High-dimensional issues}

\subsubsection{Explosion of the dimension of symptoms distributions}\label{highdim}

We are able to estimate the symptom combination distribution \big($\mathcal{P}$ of formula \eqref{vecteurinconnu}\big) of each disease $D$ provided that we can store this vector (i.e $K$ is small enough). Note that we actually need a larger vector, as our algorithm processes the information collected by the physician \textit{sequentially}. For example, we will need the $P[S_1,S_2\mid D]$ probability, which is not in $\mathcal{P}$ if $K\neq2$.
There are two possible solution for a disease with $K$ typical symptoms: i) to store the bigger vector of dimension $3^K+K$ since we would need to code in ternary to include the information "not seen yet" relative to a given symptom; or ii) to store the smaller vector of dimension $2^K+K$ and compute, on the fly, to recover desired symptoms combination probabilities from available ones.  
As we will intensively use our environment model for training our AI, we should prefer the first solution, as much as possible. 

However, it  clearly appears that we will not be able to compute/store the distribution of symptoms combination for all the diseases. Indeed when a diseases has a large number of $K$ symptoms, the dimension of the vector $\mathcal{P}^{new}$ we aim to estimate explode: $3^K+K$. 

To cope with this issue, we will use the available ontological information about symptoms, i.e the fact that a symptom can be described at different level of precision and make less stringent assumption about the dependence between symptoms (see section \ref{hypdim}).

\subsection{Relaxing the model to face potential database default}
\label{databasedefault}
So far, our model relies heavily on the assumption that expert data gives an exhaustive representation of each diseases. If a symptom has been forgotten for a disease in our expert data list, we would not be able to recover the disease.  

That is the reason why we make the assumption that a non-typical symptom (i.e. a symptom that have not been associated to the disease in the expert data) may be observed in a patient with disease $D$, but with a small ($10^{-5}$) probability and independently of other symptoms.

\section{Integrating the ontological information}
\label{ontology}

\subsection{Why we need to be concerned about the ontology of symptoms}
So far we described the diseases as combinations, more or less plausible, of symptoms. We designed algorithms inquiring about symptoms so as to find the right disease while minimizing the average number of questions to ask. We have seen that in order to learn a good strategy, we need to learn a model of the environment, i.e to learn the symptom combination distributions given the disease.    

\begin{figure}
\centering
\includegraphics[scale=0.6]{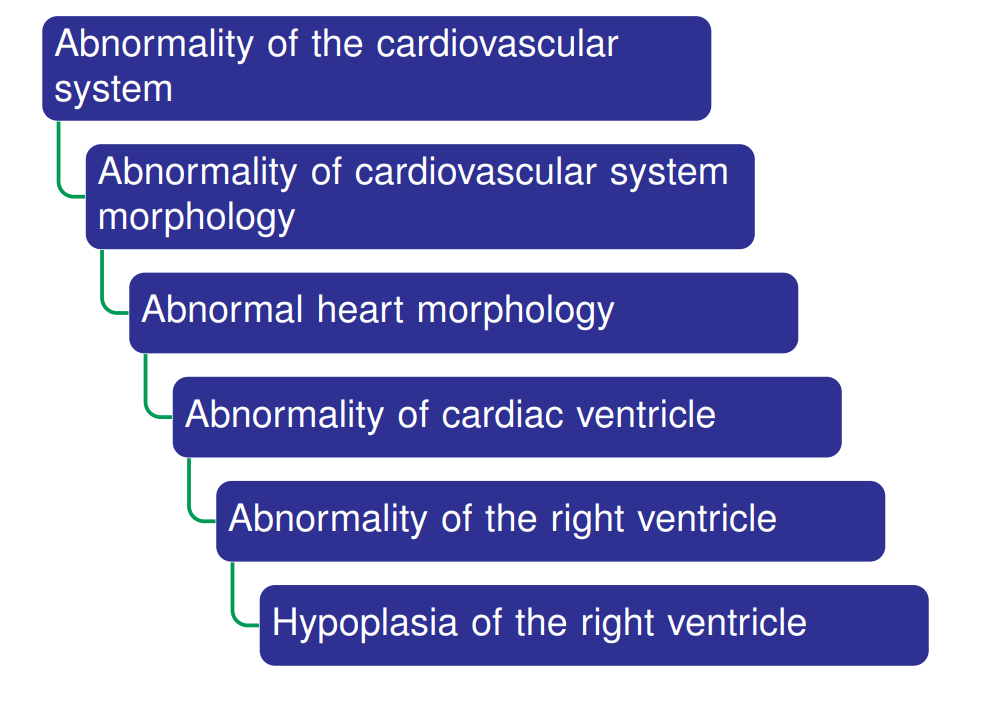}
\caption{A (very) short extract of the cardiac system ontology.}
\label{ontologygraph}
\end{figure}

Nevertheless a decision support tool built in this way will suffer from several issues. Indeed, the symptoms in medicine can be described at several level of precision. A concrete example for the abnormality "hypoplasia of the right ventricle" is displayed in the figure \ref{ontologygraph} (the terms range from the least precise to the most precise).   

In medicine, these kind of trees, called ontology, are commonly used to represent the knowledge on symptoms hierarchy.
These ontologies are built in order to capture the structure and relations between symptoms. Medical ontologies can bear an (almost) infinite level of precision. 

A naive decision support tool, i.e which would not include the available ontological information, could ask irrelevant question to a physician. For example, it is perfectly possible that our decision support would advice looking for an hypoplasia of the right ventricle when the physician already mentioned that there is no morphological abnormality of the heart. It is this kind of non-sense that we aim to solve in this section. 

Furthermore our decision support tool appears, at the moment, too rigid. Indeed, we could ask for an "hypoplasia of the right ventricle" when the physician could not give us such a precise information but rather a more imprecise one like "there is an abnormality of the cardiac ventricle". We should be able to deal with such imprecise answers and then give to the physician more freedom when interacting with our decision support tool while avoiding an explosion in computing time. 
Once again the use of the ontology will allow us such an improvement. 

\subsection{A less rigid decision support tool without computation explosion}

Each symptom of our initial database 
has been mapped to the HPO database. We, then, have been able to extract the underlying tree structure linking the different HPO codes. To be more precise, we know for each HPO code (a given description of a symptom), all its descendants (more precise description of such symptom) as such as all its ascendants (less precise description of such symptom).

\subsubsection{The idea}
As previously explained, we aim at giving more freedom to the users when describing the symptoms they observed, giving them the possibility to describe the symptoms at different level of precision.

Then instead of giving answers at a given level of precision (our initial list of symptom), we now allow the physician to choose any of the HPO code. It involves an explosion in the number of possible symptoms: our former list of symptoms references some $200$ signs when the HPO ontology has around $1300$. Both our way to modelize the symptoms combination distributions and our learning algorithms will not be able to cope with such an explosion.

Theoretically each patient could be unique if its symptoms are described to a sufficient level of accuracy. Nevertheless, when we list the typical symptoms of a disease, we try to generalize and find patterns in patient profiles. Then the idea is too still modelize the symptoms combination distributions with our initial database (the one with $200$ symptoms) preserving the ability of generalization our algorithms. We will still propose symptoms to check at the level of precision of the initial database but allow the user to give answers at a different level of precision (any HPO code can be chosen). By proceeding in this way we obtain a less rigid decision support without computation explosion since all computation are done at the initial level of precision.   

For such an objective we then need a function translating the received imprecise information (the HPO code) into usable information (presence/absence of symptoms at our precision level). Such a function will involve deterministic and stochastic rules.

\paragraph{Deterministic rules\\}


Our function associating to each HPO code the usable information associated implies some automatic (deterministic) rules. Namely:
\begin{itemize}
\item If we received a positive answer for a given HPO code, all its ascendants should be given a positive answer too. 
\item If we received a negative answer for a given HPO code, all its descendants should be given a negative answer too.
\end{itemize}
In practice we store during the medical examination all the information given about the HPO codes selected by the user. In order to compute the probability of each disease we need to check, for each HPO code and each disease, if this HPO code is in the list of the symptoms related to this disease. If not we have to check whereas ascendants or descendants of this HPO code are in this given list of symptom. Following our two deterministic rules, if the HPO code was declared to be present we have to check if ascendants are in the list, if it was declared to be absent we have to check the descendants. 

If the HPO code verifies all the following assertions it can be considered as non typical and treated in consequence (namely its presence is unlikely as in section \ref{databasedefault}):

\begin{itemize}
\item The HPO code is not in the list of symptoms related to the disease.
\item It is present and its ascendants/descendants are not in the list or it is absent and its ascendants are not in the list.
\end{itemize}

Note that the second point involves a relation that we have not studied until then. Indeed, what happens if we observed the presence of an abnormality which HPO code is not in the list of symptoms of the disease but has descendants which are in the list? This issue is studied in the next section.

\paragraph{Stochastic rules\\}

Let us assume that we have observed the presence of an "abnormal heart morphology" but that the disease we are interested in only has in its list of typical symptom the "Hypoplasia of the right ventricle". How to take into account such an imprecise information? We need stochastic rules for this issue.

When receiving the information of the presence of a HPO code, we have to determine which of its descendants are in the list of symptoms of our first database (the one which we use to build our environment model). All these symptoms have a known probability of apparition (given what we already observed) and we are able to compute them.

Indeed let us denote $L$ a list of symptom for which there is no descendant in our initial database or which are absent. Then let assume that we observed the presence of a symptom which potential descendants are $S_1^{(1)}$, $S_2^{(1)}$, $S_3^{(1)}$ and $S_4^{(1)}$ and the presence of a second symptom which potential descendants are $S_1^{(2)}$ and $S_2^{(2)}$.

There are then $4\times 2=8$ combinations possible. Indeed, without any additional assumption the number of possible combinations could be large. This is why we assume that for each imprecise answer there is only one descendant which is present at a time.  
 Our function will first compute $\forall i, j, D$:
$$P[S_i^{(1)},S_j^{(2)},L\mid D].$$

It is just the matter of searching for each $D$ which are the typical symptoms in the list $S_i^{(1)},S_j^{(2)},L$ and use the deterministic rules if necessary.

We can then compute $$P[S_i^{(1)},S_j^{(2)}\mid L]\propto P[S_i^{(1)},S_j^{(2)},L]=\displaystyle\sum_{D} P[S_i^{(1)},S_j^{(2)},L\mid D]P[D].$$

We can then display the probability of each disease (we denote $\tilde{S}$ for the fuzzy state associated to the $8$ possible states $S_i^{(1)}$ and $S_j^{(2)}$): 

$$P[D\mid \tilde{S},L]=P[D\mid L, \cup_{i,j}(S_i^{(1)}\cap S_j^{(2)})]\propto P[L,\cup_{i,j}(S_i^{(1)}\cap S_j^{(2)})\mid D] P[D]=\sum_{i,j}P[L,S_i^{(1)}, S_j^{(2)}\mid D] P[D]$$

\subsubsection{Optimize the strategy on the leaves of the ontological tree and then go back up}

Our stochastic rule can be can be expensive in terms of computing resources while it is of crucial importance for us to be able to interact quickly with our environment when training our agent. Therefore, the idea is to optimize the subtasks which start from symptoms which does not have any descendants in our database (the leaves). By this way we will not have to use our stochastic rule while training the neural networks. 

It is moreover easy to derive the strategy we have to follow when we receive an imprecise answer during an exam. We denote $\tilde{s}$ the fuzzy state and $(s^{(i)})_{i=1,...,\tilde{d}}$ the associated possible states. We can compute the Q-values in this fuzzy state by averaging on the Q-values on the possible states:

\begin{equation}
\label{average}
Q_{\pi}(\tilde{s},a)=\displaystyle\sum_{i=1}^{\tilde{d}} p(s^{(i)}\mid \tilde{s}) \times Q_{\pi}(s^{(i)},a)
\end{equation}

In practice, when receiving an imprecise answer, our algorithm should ask all the time to the physician if he could furnish a more precise answer. If not, a computation as \eqref{average} will have to be performed in real time during the examination. This computation should not last more than a second, otherwise we can consider that the provided information was not precise enough and can be overlooked.

To avoid using the stochastic rules while training our agent we will need also to remove all the action which has descendants and replace them by their leaves. In a future work it would be interesting to allow different levels of precision for the action that would suggest the neural network.

\subsection{Relations between the ontology and the symptom combination representation} \label{hypdim}

We insisted in section \ref{highdim} that there are some cases where we are not able to compute $\mathcal{P}^{new}$ but we still want to be able to compute quickly the probabilities $P[S_1,...,S_j\mid D]$ without making the assumption of conditional independence. The only solution is to relax our model of dependence between symptoms. We assumed so far that there was dependence between all the symptoms of a disease, we should now consider dependence with a less stringent approach. For the clarity of our presentation, we will consider here a two-stage deep ontology with a deeper stage for specific symptoms description and a more vague level for organs.


Let's assume we are interested in a disease with $K_1$ cardiac typical symptoms ($C_1$,...,$C_{K_1}$) and $K_2$ renal typical symptoms ($R_1$,...,$R_{K_2}$).
We denote: $$R=\begin{cases}
1 &\text{if there is at least one renal abnormalities}\\
0 &\text{otherwise} \,.
\end{cases}$$
Then we assume (precise) symptoms from distinct organs are conditionally independent given which organs have abnormalities, so we have the following decomposition: 
\begin{align*}
P[C_1,...,C_{K_1}, R_1,...,R_{K_2}\mid D]&=P[C_1,...,C_{K_1}\mid C, D]\\
&\times P[R_1,...,R_{K_2}\mid R, D]\\
&\times P[C, R \mid D].
\end{align*}
Note that even if we have lost the possibility to store dependence between precise symptoms from different organs ($C_i$ and $R_j$), we keep a model of dependence at the higher level in ontology: dependence between organs abnormalities ($C$ and $R$).

Instead of computing and storing all symptoms combinations we will just store symptoms combinations inside organs and organs combinations.

The probability of symptom combinations (i.e $P[C_1,...,C_{K_1}\mid C, D]$ in our example) will be computed solving the optimization problem~\eqref{myeq3} of section~\ref{optimproblem} with assumption to present at least one symptom (which was yet an assumption before).
The organs abnormality combinations $P[R,C\mid D]$ are computed too using~\eqref{myeq3}. When marginals $P[R\mid D]$ or $P[C \mid D]$ are not known we can treat them as missing values or try to approximate them using marginals of the lower level, temporarily making  some kind of conditional independence assumption. 

Each symptom combination can be easily computed using the law of total probability, for example we have the following decomposition:
\begin{align*}
P[\bar{R_1}, C_1 \mid D]=&P[\bar{R_1} \mid \bar{R}, D]\times P[C_1 \mid C, D]\times P[\bar{R}, C \mid D]\\
+& P[\bar{R_1} \mid R, D]\times P[C_1 \mid C, D]\times P[R, C \mid D]
\end{align*}
where $P[\bar{R_1} \mid \bar{R}, D]=1$ and all the other probabilities have been stored making these kind of computations very cheap.

This approach is in fact perfectly adapted for several diseases which manifested themselves in combinations of symptoms coming from specific organs. For example VACTERL syndrome is a rare genetic diseases defined by a combination of at least three abnormalities from three distinct organs \cite{Solomon20} among vertebral anomalies, anorectal malformation, cardiovascular anomalies, tracheoesophageal fistula, esophageal atresia, renal and/or radial anomalies and limb defects (thus defining the acronym of the disease by their first letter).  

Then we will be able to cope with any symptom combination distribution even when the number of related symptoms to a disease is high. In such a case we will have to find ascendants common to several of these symptoms (which is always possible by definition) that will organize the symptoms in groups (the organs in our example of VACTERL). We will then make the conditional independence assumption between symptoms given the ascendants.  

\begin{figure}
   \begin{minipage}[c]{.475\linewidth}     \includegraphics[scale=0.675,width=6.75cm,height=6.75cm]{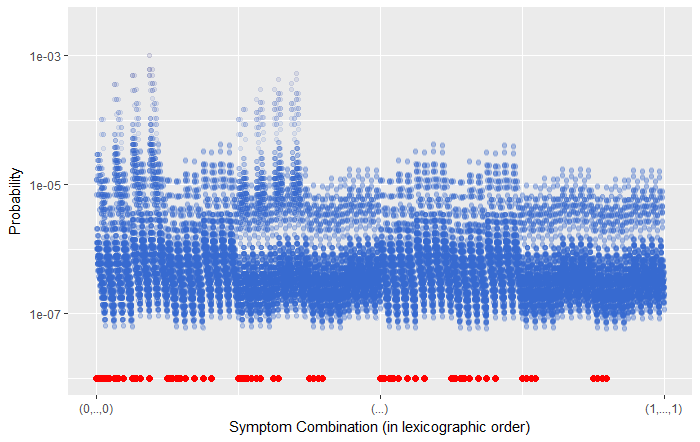}
    \caption{Symptoms combinations distribution for the VACTERL syndrome obtained by maxent with group modeling.}
      \label{vactergroup}
   \end{minipage} \hfill
   \begin{minipage}[c]{.475\linewidth}     \includegraphics[scale=0.65,width=6.50cm,height=6.50cm]{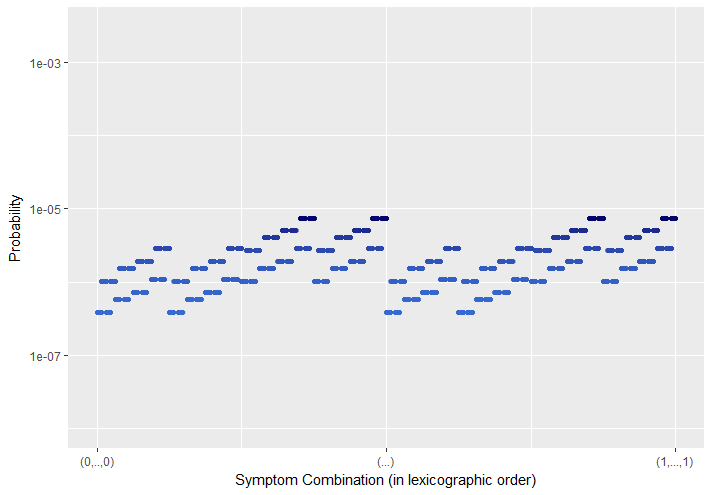}
    \caption{Symptoms combinations distribution for the VACTERL syndrome obtained by making the conditional independence assumption.}
      \label{vacterindep}
   \end{minipage}
\end{figure}

Figure~\ref{vactergroup} displays the distribution of  symptom combinations of VACTERL syndrom obtained with this group modeling. We plotted the $2^{19}\approx 500 000$ points with some in transparency for visibility reasons. Symptom combinations modeled as impossible because there are not sufficient groups turned on, are plotted in red. Comparing this plot to the one obtained making the conditional independence assumption (see Figure~\ref{vacterindep}), it appears that group modeling adds much information about the distribution of symptom combinations. The visible symmetries of the distribution are only due to the lexicographic order we use for the different symptom combinations. 

\section{Conclusion}

We have presented in this work a novel notion, as far as we know, of what should be a good decision support tool for a rare disease diagnostic task. We took into account the need, in medicine, to achieve a high level of certainty when possessing a diagnostic. We try to minimize the average number of medical tests to be performed before reaching this level of certainty. We investigated several reinforcement learning algorithms and make them operable in our high-dimensional and reward-sparse setting. To do this we broke the initial task into several sub-tasks and learned a policy for each sub-tasks. We proved that an appropriate use of the intersections between the sub-tasks can significantly accelerate the learning procedure. 

Furthermore we reconnected with the first works on expert systems which used probabilistic reasoning. This is due to our target application, we are interested in rare diseases and then we can not work without expert knowledge which is generally expressed as conditional probabilities. We presented a way to integrate expert knowledge with clinical data processed by the decision support tool.    

Finally we showed that it is possible to integrate the ontological information while remaining in our probabilistic setting. This result in a less rigid decision support tool without computation explosion.         

\bibliographystyle{plain}
\bibliography{biblio}

\end{document}